\documentclass[conference]{IEEEtran}
\IEEEoverridecommandlockouts
\usepackage{float}  
\usepackage{booktabs}
\usepackage{amssymb}
\usepackage{tabularray}
\usepackage{multirow}
\usepackage{graphicx}
\usepackage{xspace}
\usepackage{setspace}
\usepackage{amsmath}
\usepackage{algorithm}
\usepackage{algorithmic}
\newcommand{\framework}{\textnormal{FedS2R}\xspace}
\usepackage{cite}
\usepackage{amsmath,amssymb,amsfonts}
\usepackage{algorithmic}
\usepackage{graphicx}
\usepackage{textcomp}
\usepackage{xcolor}
\def\BibTeX{{\rm B\kern-.05em{\sc i\kern-.025em b}\kern-.08em
    T\kern-.1667em\lower.7ex\hbox{E}\kern-.125emX}}

\begin{document}
\title{FedS2R: One-Shot Federated Domain Generalization for Synthetic-to-Real Semantic Segmentation in Autonomous Driving}
\author{
    \IEEEauthorblockN{Tao Lian\thanks{This work was supported by the Spanish grant Ref. PID2024-157936NB-I00 (HAMILTON project), funded by MICIU/AEI/10.13039/501100011033 and by FEDER, EU. A. M. López acknowledges financial support for his general research activities from ICREA under the ICREA Academia Program. T. Lian acknowledges support for his research from the China Scholarship Council through the Ph.D. Grant Ref. 202102890072. All authors acknowledge the support of the Generalitat de Catalunya through the CERCA Program and its ACCIÓ Agency for CVC’s general activities.}}
    \IEEEauthorblockA{\textit{Computer Vision Center (CVC)} \\
    \textit{Univ. Autònoma de Barcelona (UAB)}\\
    Barcelona, Spain \\
    tlian@cvc.uab.cat}
    \and
    \IEEEauthorblockN{Jose L. G\'{o}mez}
    \IEEEauthorblockA{\textit{Computer Vision Center (CVC)} \\
    \textit{Univ. Autònoma de Barcelona (UAB)}\\
    Barcelona, Spain \\
    jlgomez@cvc.uab.cat}
    \and
    \IEEEauthorblockN{Antonio M. L\'opez}
    \IEEEauthorblockA{\textit{Computer Vision Center (CVC)} \\
    \textit{Univ. Autònoma de Barcelona (UAB)}\\
    Barcelona, Spain \\
    antonio@cvc.uab.cat}
}
\maketitle
\begin{abstract}
Federated domain generalization has shown promising progress in image classification by enabling collaborative training across multiple clients without sharing raw data. However, its potential in the semantic segmentation of autonomous driving remains underexplored. In this paper, we propose \framework, the first one-shot federated domain generalization framework for synthetic-to-real semantic segmentation in autonomous driving. \framework comprises two components: an inconsistency-driven data augmentation strategy that generates images for unstable classes, and a multi-client knowledge distillation scheme with feature fusion that distills a global model from multiple client models. Experiments on five real-world datasets, Cityscapes, BDD100K, Mapillary, IDD, and ACDC, show that the global model significantly outperforms individual client models and is only 2 mIoU points behind the model trained with simultaneous access to all client data. These results demonstrate the effectiveness of FedS2R in synthetic-to-real semantic segmentation for autonomous driving under the federated learning setting.
\end{abstract}

\section{Introduction}
\label{sec:intro}
Semantic segmentation \cite{csurka2022semantic} is a key perception task in autonomous driving \cite{janai2020computer,tampuu2020survey}, enabling pixel-wise understanding of road scenes. Training such models typically requires large-scale datasets with precise pixel-level annotations, which are expensive and labor-intensive to obtain. To avoid manual labeling, many studies leverage synthetic datasets with automatically generated pixel-level annotations. Despite this advantage, models trained solely on synthetic data often fail to generalize well to real-world domains due to domain gaps. To mitigate this problem, several works combine multiple synthetic datasets \cite{lian2024divide,gomez2025all,gomez2023co}, increasing scenario diversity and reducing domain shift. Nevertheless, widely used synthetic datasets \cite{gomez2025all, richter:2016GTA5, ros:2016, wrenninge:2018synscapes} are typically built with proprietary engines and assets, which can limit accessibility and usability.

With the development of generative AI tools, synthetic data creation has become more accessible. However, obtaining unrestricted access to labeled datasets for model training is often infeasible. Dataset providers often enforce strict licensing terms that prohibit redistribution or unauthorized usage, particularly in commercial or sensitive domains. Large-scale datasets also require significant storage and infrastructure resources. These constraints motivate the leveraging of existing models as carriers of knowledge, rather than relying solely on datasets. Sharing models is generally more feasible than sharing raw data, especially in scenarios such as autonomous driving companies with global branches where cross-border data exchange is restricted; collaborative research across institutions with incompatible data-management policies; and publicly released models where the underlying datasets cannot be disclosed due to licensing or proprietary concerns. Federated learning \cite{feddrive,fedcdg,elcfs,ccst,ga,asam} provides a natural solution by keeping data local while allowing only model weights or gradients to be shared with a central server. This paradigm preserves privacy, satisfies legal and ethical requirements, and enables learning of domain-invariant representations. Existing work, however, focuses primarily on image classification \cite{fedavg,fedcdg,ccst,ga,FEDCVAE-KD,FEDDEO} and typically assumes multiple rounds of communication, which is impractical when model providers cannot participate in iterative updates. Recent advances in one-shot federated learning \cite{zhang2022dense,FEDDEO,FEDCVAE-KD} reduce communication to a single round, making them more compatible with such settings. Nonetheless, these methods focus on classification, and applying them to semantic segmentation, which is essential for autonomous driving, has not been thoroughly investigated.
\begin{figure}[t]
    \centering
    \includegraphics[width=1\linewidth]{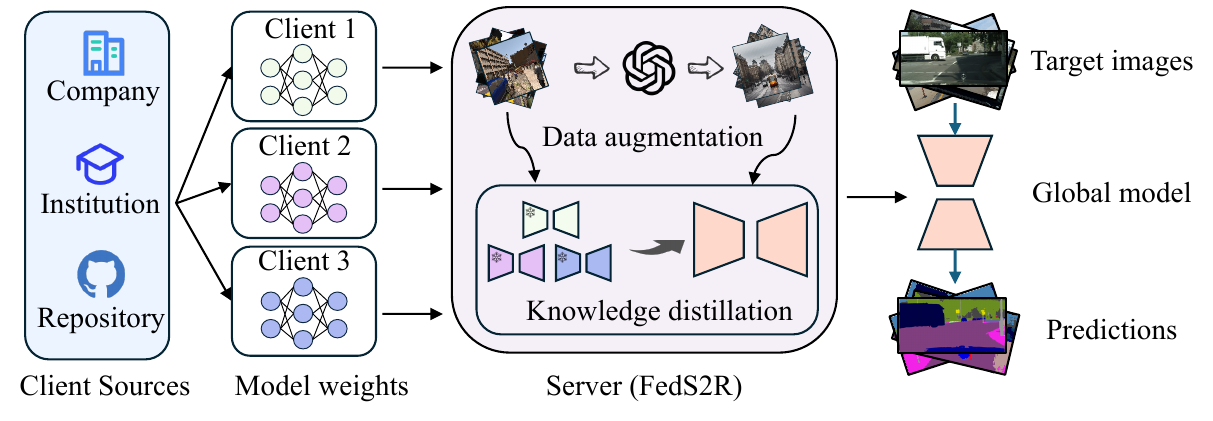}
    \caption{Our framework is designed for scenarios where clients (e.g., companies, institutions, or repositories providing model weights) train models on their respective local datasets and send the resulting weights to a central server. The server then utilizes its private dataset, together with the received client models, to train a global model. The proposed \framework performed on the server consists of two stages: (a) data augmentation using ChatGPT and diffusion models to generate images; and (b) knowledge distillation to distill multiple client models into a global model. The star symbol (\texttt{*}) indicates that the corresponding model weights are frozen.}
    \label{fig:intro}
\end{figure}

In this paper, we propose \framework, a novel one-shot federated domain generalization framework for semantic segmentation from synthetic to real domains. As shown in Fig. \ref{fig:intro}, our framework is designed for federated scenarios in which multiple clients (e.g., companies, research institutions, or repositories) independently train models on their respective local datasets. Each client retains its data privately and only sends model weights to a central server, rather than the raw data. Upon receiving the client models, the central server aggregates and leverages them, along with its own private dataset, referred to as the server dataset, to train a more generalized global model. \framework consists of two stages: inconsistency-driven data augmentation and multi-client knowledge distillation. In the distillation stage, \framework leverages knowledge distillation \cite{hinton2015distilling,liu2019structured,liu2019knowledge} to aggregate multiple client models into a global model without accessing any client data. Moreover, the global model is trained using a distillation dataset (server dataset) in a fully unsupervised manner, relying only on the pseudo labels provided by the client models on this dataset. Note that this dataset requires no annotation, providing flexibility and adaptability. In our scenario, we limit ourselves to synthetic data due to the synth-to-real domain gap challenge, but real-world data could be easily added. We adopt the state-of-the-art Mask2Former architecture \cite{cheng2022masked}, known for its robust performance in segmentation tasks, and assume each client provides models with this architecture. Following the aforementioned example of a company with facilities worldwide, our approach mimics a situation where all divisions use the same state-of-the-art model to share weights. To the best of our knowledge, this is the first federated domain generalization framework designed for synthetic-to-real semantic segmentation in autonomous driving. It is also the first to distill multiple Mask2Former models into a single global model without requiring ground-truth for the distillation data. Experiments on five real-world datasets, Cityscapes, BDD100K \cite{yu:2020BDD}, Mapillary \cite{neuhold:2017mapillary}, IDD \cite{varma2019idd}, and ACDC \cite{sakaridis2021acdc}, demonstrate that the distilled global model outperforms each client model individually and is only ~2 mIoU points behind a trained global model with simultaneous access to all the data.  Marking a significant contribution in federated domain generalization for semantic segmentation of autonomous driving. In summary, our contributions are threefold:
\begin{itemize}
    \item We propose a novel federated learning framework for semantic segmentation from synthetic to real domains for autonomous driving.
    \item We introduce transformer-based Mask2Former models into the federated domain generalization setting, demonstrating their feasibility and advantages.
    \item We achieve consistent generalization performance of the global model on multiple real-world datasets without requiring any ground-truth.
\end{itemize}

\section{Related work}
\label{sec:formatting}
\textbf{Federated Learning}. Federated learning is a distributed client-server paradigm that enables multiple clients to collaboratively train a global model without sharing data. The most widely used algorithm, FedAvg \cite{fedavg}, performs multiple communication rounds by aggregating client models on a central server and redistributing the global model back to the clients for further training. Subsequent works build on FedAvg, further improving the performance of image classification \cite{fedcdg,ccst,ga} and medical image segmentation \cite{elcfs}. In contrast, relatively fewer studies have focused on semantic segmentation. FedSeg \cite{miao2023fedseg} proposes a modified cross-entropy loss to address the foreground-background inconsistency caused by class heterogeneity in federated segmentation. ASAM \cite{asam} improves generalization by applying sharpness-aware minimization locally and averaging stochastic weights on the server to narrow the gap between federated and centralized models. FMTDA \cite{FMTDA} tackles both server-client and inter-client domain shifts by formulating a domain adaptation problem with one source and multiple target domains. Despite their contributions, these methods require multiple communication rounds and retraining on the client side, introduce significant communication overhead, client dependency, and deployment complexity. 

To reduce communication overhead and client dependency, one-shot federated learning has emerged as an alternative, requiring only a single round of weight upload from clients. DENSE \cite{zhang2022dense} introduces a two-stage framework involving data generation and model distillation to train a global model without accessing client data or performing client retraining. FedDEO \cite{FEDDEO} learns client-specific descriptions to guide diffusion models in generating synthetic data to train the global model. FedCVAE \cite{FEDCVAE-KD} trains conditional VAEs on each client, then leverages the uploaded decoders and label distributions on the server side to generate synthetic samples for training the global model. Although these one-shot methods effectively reduce communication costs via knowledge distillation, they primarily target image classification. In this work, we explore a one-shot federated setting on semantic segmentation, especially relevant for the autonomous driving community due to concerns about data privacy.

\textbf{Federated Domain Generalization}. Federated domain generalization integrates federated learning and domain generalization to train models across decentralized domains without data sharing. This paradigm is particularly well-suited for scenarios where data centralization is infeasible due to privacy, legal, or storage constraints. Although federated domain generalization has gained attention in image classification \cite{fedcdg,ccst,ga,asam}, semantic segmentation remains underexplored. FedDrive \cite{feddrive} introduces a benchmark for semantic segmentation under federated domain generalization. However, it evaluates performance across different cities within the Cityscapes dataset, undermining the domain generalization. It also employs a lightweight model and the standard FedAvg communication strategy, which limits its applicability in real-world scenarios where more powerful architectures and one-shot communication settings are typically required.
Furthermore, in synthetic-to-real domains in federated learning, DDI \cite{ddi} constructs clients with mixed distributions from GTA5 and Cityscapes to study a synthetic-plus-real to real setting. Moreover, most segmentation-oriented federated learning frameworks \cite{feddrive,miao2023fedseg} assume homogeneous data between clients (e.g., partitions of the same dataset), whereas real-world federated setups often involve heterogeneous domains. This discrepancy is especially pronounced in synthetic-to-real scenarios, where client models trained on different synthetic datasets face significant domain gaps. Our work addresses federated domain generalization for semantic segmentation under a synthetic-to-real setting, where each client is trained on a separate synthetic dataset.
\section{Method}
\label{method}
\begin{figure*}[t]
  \centering
  \includegraphics[width=1\textwidth]{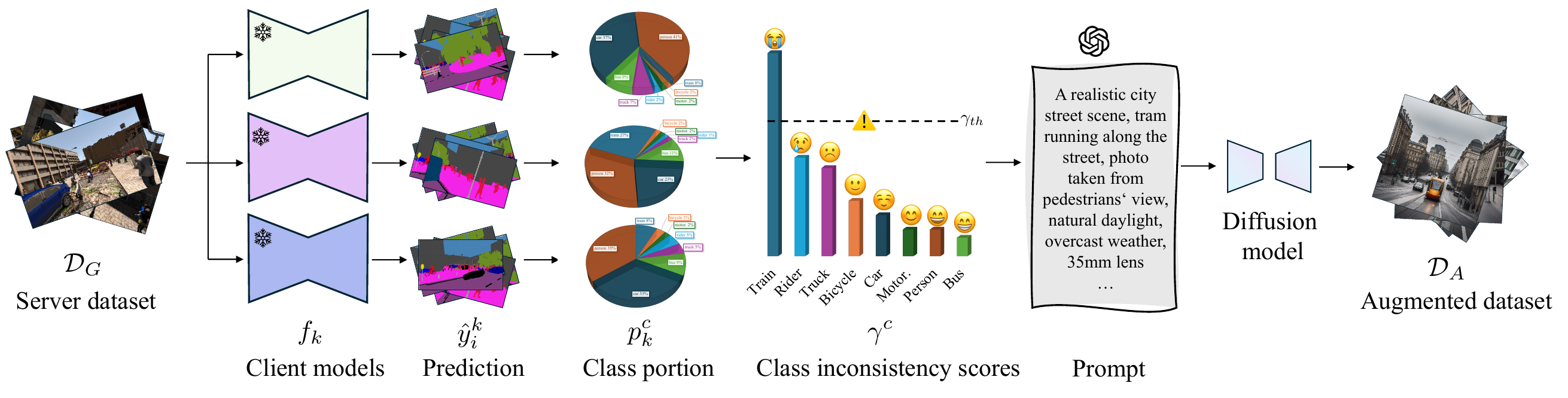}
   \caption{Overview of the inconsistency-driven data augmentation strategy. The strategy first employs client models to make predictions on the server dataset and calculates class-level inconsistency scores based on the class portions to identify unstable classes. Subsequently, prompts related to these unstable classes are generated using ChatGPT, and a diffusion model is used to synthesize corresponding images. The star symbol (\texttt{*}) indicates that the corresponding model weights are frozen.}
   \label{incon}
\end{figure*}
In this section, we present our framework, $\framework$, which comprises two main components: (1) inconsistency-driven data augmentation to identify and synthesize unstable classes, and (2) multi-client knowledge distillation with feature-level alignment and class-mask-level loss to train a robust global model. Prior studies \cite{fedavg,feddrive,ga,ccst} generally employ lightweight architectures, which result in constrained representational capacity and suboptimal performance for the semantic segmentation task. Thus, we adopt the high-capacity Mask2Former for both clients and the server to capture complex scene semantics more accurately. Moreover, $\framework$ operates in a one-shot federated setting, where each client uploads its model weights only once, making it more practical for real-world deployment in privacy-sensitive applications.
\subsection{Problem Formulation}
Assume there are $K$ clients and a server. Each client holds a private dataset $\mathcal{D}_K=\{x^i_k,y^i_k\}_{i=1}^{N_K}$ covering $C$ classes, while the server has no access to any raw client data. Each client trains a local segmentation model $f_k$ on its dataset and uploads the model weights to the server. The server maintains a separate dataset  $\mathcal{D}_G = \{x_i\}_{i=1}^{N_G}$ as server dataset. The objective is to train a global model $f_G$ to predict the same $C$ classes and outperform individual client models on unseen real-world domains, while adhering to data privacy constraints and without requiring access to target domain data. Furthermore, $f_G$ should be as close as possible to a global model trained with unrestricted access to all the data, which serves as our upper bound.
\begin{algorithm}[t]
\renewcommand{\algorithmicrequire}{\textbf{Input:}}
\renewcommand{\algorithmicensure}{\textbf{Output:}}
\caption{Inconsistency-Driven Augmentation}
\label{alg:ida}
\begin{algorithmic}[1]
\REQUIRE Server dataset $\mathcal{D}_G$, client models $\{f_k\}_{k=1}^K$, dynamic foreground class set $\mathcal{M}$, threshold $\gamma_{\text{th}}$
\ENSURE Augmented dataset $\mathcal{D}_A$
\STATE Initialize matrix $\mathbf{R} \in \mathbb{R}^{K \times |\mathcal{M}|}$
\FOR {each image $x_i \in \mathcal{D}_G$}
    \FOR {each client $k$}
        \STATE Predict pseudo-label $\hat{y}_i^k = f_k(x_i)$
        \STATE Initialize class count vector $N_k \in \mathbb{R}^{|\mathcal{M}|}$
        \FOR {each class $c \in \mathcal{M}$}
            \STATE Count $N_k^c \leftarrow \text{pixels in } \hat{y}_i^k \text{ labeled } c$
        \ENDFOR
        \STATE Normalize and update $\mathbf{R}[k, :] \leftarrow \text{Softmax}(N_k)$
    \ENDFOR
\ENDFOR
\STATE Compute mean $\mu^c$, std $\sigma^c$, and inconsistency score $\gamma^c$
\STATE Select $\mathcal{R} = \{ c \mid \gamma^c > \gamma_{\text{th}} \}$
\FOR {each class $c \in \mathcal{R}$}
    \STATE Generate scene-aware prompt $p_c$ via LLM
    \FOR {$n = 1$ to $N$}
        \STATE $\mathcal{D}_A \leftarrow \text{Diffusion}(p_c)$
    \ENDFOR
\STATE Construct distillation set: $\mathcal{D}_{GA} \gets \mathcal{D}_G \cup \mathcal{D}_A$
\ENDFOR
\end{algorithmic}
\end{algorithm}
\subsection{Inconsistency-driven Data Augmentation}
\label{sec:inconsistency_aug}
We assume each client model is independently trained on a different dataset. Due to varying class distributions across client datasets, client models often yield inconsistent predictions on the same image, particularly for unbalanced or inconsistent classes. Moreover, the server dataset $\mathcal{D}_G$ may also exhibit class imbalance, which can lead to instability in training the global model. Both issues can hinder the effectiveness of the distilled global model. 

To address this issue, we introduce an \textit{inconsistency-driven data augmentation} strategy that quantifies class-level prediction inconsistency to identify unstable classes and employs a diffusion model to generate additional images for these classes. The detailed procedure is illustrated in Figs. \ref{incon} and implemented in Algorithm \ref{alg:ida}. First, we use client models $f_k$ to make predictions $\hat{y}_i^k \in {[ 0, 1, \dots, C-1 ]}$ for images $x_i$ in $\mathcal{D}_G$. Since the most relevant objects for autonomous driving that normally suffer from imbalance are the dynamic ones, we select a subset with dynamic foreground classes $\mathcal{M} \subseteq {[0, 1, \dots, C-1]}$ to be considered in the inconsistency analysis. 
For each client $k$ and class $c \in \mathcal{M}$, we calculate the relative class proportion as $p_k^c = N_k^c / \sum_{c \in \mathcal{M}} N_k^{c}$, where $N_k^c$ denotes the number of pixels predicted as class $c$ by the client $k$ across all images in $\mathcal{D}_G$.
This proportion reflects the relative frequency of class $c$ among class set $\mathcal{M}$. To quantify class inconsistency scores across clients, we compute the mean and standard deviation of each class proportion as follows:
\begin{equation}
\begin{aligned}
\mu^c &= \frac{1}{K} \sum_{k=1}^K p_k^c , \quad
\sigma^c = \sqrt{ \frac{1}{K} \sum_{k=1}^K (p_k^c - \mu^c)^2 }.
\end{aligned}
\end{equation}
We then define the inconsistency score as $\gamma^c = \sigma^c / (\mu^c + \epsilon)$, where $\epsilon$ is a small constant to ensure numerical stability.
A higher $\gamma^c$ indicates either a low class presence or high disagreement across clients, suggesting this class is unstable. Based on this score, we define the unstable class set $\mathcal{R}$ as $\mathcal{R} = \left\{ c \in \mathcal{M} \mid \gamma^c > \gamma_{\text{th}} \right\}$,
where $\gamma_{\text{th}}$ is a threshold. For each class $c \in \mathcal{R}$, we use a large language model (e.g., ChatGPT) to generate semantically descriptive prompts tailored to autonomous driving scenes. These prompts are fed into a pre-trained diffusion model to synthesize diverse and photorealistic images containing class $c$. The distillation dataset $\mathcal{D}_{GA}$ is constructed by combining the server dataset $\mathcal{D}_G$ with the augmented dataset $\mathcal{D}_A$ produced by the diffusion model $\mathcal{D}_{GA} = \mathcal{D}_G \cup \mathcal{D}_A$.
This strategy enhances the representation of unstable classes in the server dataset while preserving the privacy-aware and annotation-free nature of the framework.
\subsection{Multi-client Feature Fusion and Distillation}
\begin{figure*}[t]
  \centering
  \includegraphics[width=1\linewidth]{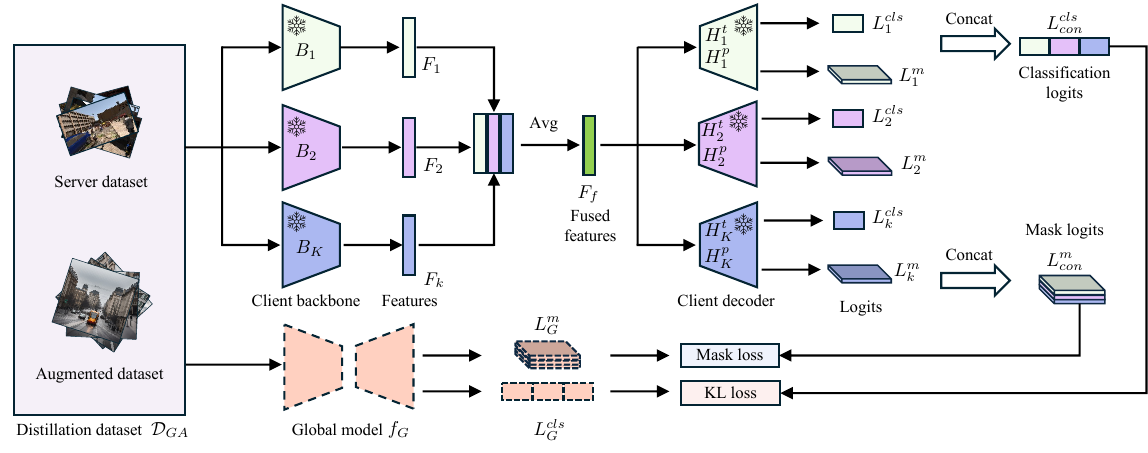}
   \caption{Overview of multi-client distillation with feature fusion. Client models extract backbone features, which are averaged and passed through the decoder to produce classification and mask logits. These logits are then respectively concatenated across clients and used to compute the distillation loss with the corresponding logits from the global model.}
   \label{multi-client}
\end{figure*}
\begin{algorithm}[t]
\renewcommand{\algorithmicrequire}{\textbf{Input:}}
\renewcommand{\algorithmicensure}{\textbf{Output:}}
\caption{Multi-client Feature Fusion and Distillation}
\label{alg:fusion_distillation}
\begin{algorithmic}[1]
\REQUIRE Distillation dataset $\mathcal{D}_{GA}$, client models $\{f_k=(B_k, H_k^t, H_k^p)\}_{k=1}^K$, untrained global model $f_G$
\ENSURE Trained global model $f_G$
\STATE Set learning rate $\eta$, loss weight $\lambda_{cls}$, $\lambda_m$
\FOR {minibatch $x \in \mathcal{D}_{GA}$}
    \FOR {$k = 1$ to $K$}
    \STATE Client backbone features: $F_k \gets B_k(x)$
    \ENDFOR
    \STATE Fuse features: $F_{f} \gets \frac{1}{K} \sum_{k=1}^K F_k$
    \FOR {$k = 1$ to $K$}
        \STATE Classification logits: $L_k^{cls} \gets H_k^t(F_{f})$
        \STATE Mask logits: $L_k^m \gets H_k^p(F_{f})$
    \ENDFOR
    \STATE Concatenate: $L_{con}^{cls} \gets Concat(L_k^{cls})$
    \STATE Concatenate: $L_{con}^m \gets Concat(L_k^m)$
    \STATE Global model logits: $L_G^{cls},L_G^m \gets f_G(x)$
    \STATE $\mathcal{L}_{cls} \gets KL(L_{con}^{cls},L_G^{cls})$
    \STATE $\mathcal{L}_m \gets BCE(L_{con}^m,L_G^m)+ Dice(L_{con}^m,L_G^m) $
    \STATE Total loss: $\mathcal{L}_{\text{total}} \gets \lambda_{\text{cls}} \cdot \mathcal{L}_{\text{cls}} + \lambda_m \cdot \mathcal{L}_m$
    \STATE Update global model: $\theta \gets \theta - \eta \cdot \nabla_\theta \mathcal{L}_{\text{total}}$
\ENDFOR
\end{algorithmic}
\end{algorithm}
\label{sec:fusion_distillation}
Following the inconsistency-driven strategy, we perform knowledge distillation to transfer information from client models $\{f_k\}_{k=1}^{K}$ to a global model $f_G$. Each client model $f_k$ follows the Mask2Former architecture and comprises a backbone $B_k$, a pixel decoder $H_k^p$, and a transformer decoder $H_k^t$. The global model $f_G$ deployed on the server adopts the same architecture. The procedure is illustrated in Fig. \ref{multi-client} and implemented in Algorithm \ref{alg:fusion_distillation}. 
Given an image $x \in \mathcal{D}_{GA}$, with $x \in \mathbb{R}^{3 \times H \times W}$, each client backbone $B_k$ extracts intermediate feature maps as $F_k = B_k(x) \in \mathbb{R}^{C_f \times H' \times W'}$
where $C_f$ denotes the number of feature channels, and $H'$, $W'$ represent the downsampled spatial dimensions. 
Since client models are trained on heterogeneous datasets, the extracted features $F_k$ may capture different semantic cues.
To integrate these diverse representations, we perform feature-level fusion by averaging features from all client backbones: $F_{f} = (1/K) \sum_{k=1}^{K} F_k \in \mathbb{R}^{C_f \times H' \times W'}$.
The fused features are then passed through the decoders of each client to produce classification logits $L_k^{cls}$ and mask logits $L_k^m$:
\begin{equation}
\begin{aligned}
L_k^{cls} &= H_k^{t}(F_{f}) \in \mathbb{R}^{Q \times (C+1)} \\
L_k^{m} &= H_k^{p}(F_{f}) \in \mathbb{R}^{Q \times H' \times W'},
\end{aligned}
\end{equation}
where $Q$ is the number of object queries and $C+1$ includes an additional background class. Although all client models are trained on datasets with the same class space $C$, the object queries in Mask2Former are randomly initialized and do not possess fixed semantic associations. Consequently, query indices are not aligned across models, i.e., query $q$ in client $A$ may attend to entirely different classes or spatial regions than the same index $q$ in client $B$. This lack of semantic correspondence makes it infeasible to directly compare or aggregate query-level logits across clients in a federated setting. Although inspired by \cite{li2024knowledge}, which relies on ground-truth labels for bipartite matching between student and teacher models, we adopt a label-free strategy by equally aggregating all query outputs, without requiring any ground-truth-based matching. Specifically, we concatenate classification logits $L_k^{cls}$ and mask logits $L_k^m$ from all clients individually:
\begin{equation}
\begin{aligned}
L_{con}^{cls} &= \mathrm{Concat}(L_1^{cls}, \dots, L_K^{cls}) \in \mathbb{R}^{KQ \times (C+1)} \\
L_{con}^m     &= \mathrm{Concat}(L_1^m, \dots, L_K^m) \in \mathbb{R}^{KQ \times H' \times W'},
\end{aligned}
\label{eq:concat}
\end{equation}
Then, the global model $f_G$ processes the same image $x$ to generate its own classification logits $L_G^{cls} \in \mathbb{R}^{KQ \times (C+1)}$ and mask logits $L_G^m \in \mathbb{R}^{KQ \times H' \times W'}$.
In Mask2Former, classification and mask logits collectively determine the final segmentation output. To effectively distill the knowledge learned by client models into the global model, we apply two types of loss: classification distillation loss and mask distillation loss.  For classification, we use temperature-scaled Kullback-Leibler (KL) divergence as classification distillation loss:
\begin{equation}
\begin{aligned}
\mathbf{p_c}^{q} &= \log \left(\mathrm{Softmax}\left( \frac{L_{con}^{cls,q}}{\tau} \right)\right) \in \mathbb{R}^C \\
\mathbf{q_c}^{q} &= \mathrm{Softmax}\left( \frac{L_{G}^{cls,q}}{\tau} \right) \in \mathbb{R}^C \\
\mathcal{L}_{cls} &= \frac{1}{KQ} \sum_{q=1}^{KQ} \sum_{c=1}^C p_c^{q} \log \left( \frac{p_c^{q}}{q_c^{q}} \right),
\end{aligned}
\label{eq:kl_loss}
\end{equation}
where $\tau$ is a temperature scaling parameter and $q$ indexes the object query. The KL divergence loss encourages the global model to align with the probabilistic output distributions (i.e., soft predictions) of the client models, enabling it to capture the class-level knowledge in their predictions. 
For the mask distillation loss $\mathcal{L}_m$, we combine binary cross-entropy (BCE) and Dice loss: $\mathcal{L}_{m} = BCE(L_G^m, L_{con}^m) + Dice(L_G^m, L_{con}^m)$. While BCE loss enforces pixel-wise consistency between global and client predictions, Dice loss emphasizes the quality of region-level overlap, which is particularly beneficial for handling small or imbalanced foreground regions.
%
The total loss used to train the global model is a weighted sum of the classification and mask losses:
\begin{equation}
\mathcal{L}_{total} = \lambda_{cls} \cdot \mathcal{L}_{cls} + \lambda_{m} \cdot \mathcal{L}_{m},
\label{eq:total_loss}
\end{equation}
where $\lambda_{cls}$ and $\lambda_{m}$ are weight coefficients for each loss.
\section{Experiments}
\begin{table*}[t]
  \caption{Performance of FedAvg and ours under various client-server dataset configurations. “Baseline” denotes a centralized setting where the server is allowed to access all client data for direct training. “GSUS” refers to the merged dataset consisting of GTA5, Synscapes, UrbanSyn, and Synthia, representing the upper bound of model performance.}
  \centering
  \setlength{\tabcolsep}{20pt}
  \resizebox{1\textwidth}{!}{
    \begin{tabular}{c|cccc|cccccc}
    \toprule
    \multirow{2}[4]{*}{Method} & \multicolumn{4}{c|}{Dataset}  & \multicolumn{6}{c}{mIoU} \\
    \cmidrule{2-11}          
          & Client 1 & Client 2 & Client 3 & Server & Cityscapes & BDD100K & Mapillary & IDD & ACDC & Average \\
    \midrule
    Baseline 
          & - & - & - & GSUS & 68.6 & 55.5 & 68.0 & 58.3 & 53.2 & 60.7 \\
    \midrule
    FedAvg & \multirow{2}[2]{*}{GTA5} & \multirow{2}[2]{*}{Synscapes} & \multirow{2}[2]{*}{Urbansyn} & \multirow{2}[2]{*}{Synthia} 
          & 45.0 & 34.9 & 42.8 & 33.3 & 30.2 & 37.2 \\
    Ours &  &  &  &  & \textbf{67.9} & \textbf{53.1} & \textbf{65.1} & \textbf{56.5} & \textbf{49.8} & \textbf{58.5} \\
    \midrule
    FedAvg & \multirow{2}[2]{*}{GTA5} & \multirow{2}[2]{*}{Synscapes} & \multirow{2}[2]{*}{Synthia} & \multirow{2}[2]{*}{Urbansyn} 
          & \textbf{67.0} & 46.2 & 62.6 & 51.7 & 42.1 & 53.9 \\
    Ours &  &  &  &  & 64.0 & \textbf{53.9} & \textbf{65.8} & \textbf{55.9} & \textbf{52.5} & \textbf{58.4} \\
    \midrule
    FedAvg & \multirow{2}[2]{*}{GTA5} & \multirow{2}[2]{*}{Synthia} & \multirow{2}[2]{*}{Urbansyn} & \multirow{2}[2]{*}{Synscapes} 
          & 56.8 & 39.8 & 56.9 & 46.5 & 42.6 & 48.5 \\
    Ours &  &  &  &  & \textbf{64.9} & \textbf{54.8} & \textbf{66.2} & \textbf{57.6} & \textbf{48.3} & \textbf{58.4} \\
    \midrule
    FedAvg & \multirow{2}[2]{*}{Synthia} & \multirow{2}[2]{*}{Synscapes} & \multirow{2}[2]{*}{Urbansyn} & \multirow{2}[2]{*}{GTA5} 
          & 64.1 & 50.2 & 63.2 & \textbf{53.4} & 47.6 & 55.7 \\
    Ours &  &  &  &  & \textbf{64.6} & \textbf{50.8} & \textbf{63.5} & 53.2 & \textbf{47.8} & \textbf{56.0} \\
    \bottomrule
    \end{tabular}%
  }
  \label{tab:comparison}
\end{table*}
\begin{table*}[t]
  \caption{Per-class IoU results on Cityscapes under different federated client-server dataset configurations. “Baseline” denotes a centralized setting where the server is allowed to access all client data for direct training. “GSUS” refers to the merged dataset consisting of GTA5, Synscapes, UrbanSyn, and Synthia, representing the upper bound of model performance.}
  \centering
  \resizebox{\textwidth}{!}{
    \begin{tabular}{c|c|ccccccccccccccccccc|c}
    \toprule
    \multirow{2}[4]{*}{Method} & \multirow{2}[4]{*}{Server} & \multicolumn{19}{c|}{IoU} &  \\
    \cmidrule{3-22}
          &       & \multicolumn{1}{c}{Road} & \multicolumn{1}{c}{S.walk} & \multicolumn{1}{c}{Build} & \multicolumn{1}{c}{Wall} & \multicolumn{1}{c}{Fence} & \multicolumn{1}{c}{Pole} & \multicolumn{1}{c}{Tr. Light} & \multicolumn{1}{c}{Tr. Sign} & \multicolumn{1}{c}{Vege.} & \multicolumn{1}{c}{Terrain} & \multicolumn{1}{c}{Sky} & \multicolumn{1}{c}{Person} & \multicolumn{1}{c}{Rider} & \multicolumn{1}{c}{Car} & \multicolumn{1}{c}{Truck} & \multicolumn{1}{c}{Bus} & \multicolumn{1}{c}{Train} & \multicolumn{1}{c}{M.cycle} & \multicolumn{1}{c|}{Bicycle} & \multicolumn{1}{c}{mIoU} \\
    \midrule
    Baseline & GSUS  & 91.7  & 58.2  & 91.1  & 55.8  & 57.6  & 63.2  & 66.0  & 74.2  & 91.4  & 53.9  & 94.5  & 80.2  & 55.0  & 93.2  & 57.9  & 65.0  & 40.6  & 42.6  & 70.5  & 68.6  \\
    \midrule
    FedAvg & \multirow{2}[2]{*}{Synthia} & 80.4 & 37.2 & 87.4 & 26.6 & 3.6 & 57.0 & 49.4 & 42.1 & 86.9 & 0.0 & 89.3 & 71.7 & 21.3 & 85.7 & 0.0 & 34.0 & 0.0 & \textbf{42.7} & 39.4 & 45.0 \\
    Ours   &                              & \textbf{92.3} & \textbf{56.6} & \textbf{91.0} & \textbf{50.5} & \textbf{56.7} & \textbf{58.3} & \textbf{58.0} & \textbf{71.2} & \textbf{90.7} & \textbf{51.9} & \textbf{92.1} & \textbf{79.5} & \textbf{56.3} & \textbf{91.9} & \textbf{58.1} & \textbf{69.8} & \textbf{56.8} & 39.8 & \textbf{68.5} & \textbf{67.9} \\
    \midrule
    FedAvg & \multirow{2}[2]{*}{Urbansyn} & 90.3 & 53.0 & 90.2 & 36.4 & \textbf{56.1} & \textbf{63.0} & \textbf{64.6} & \textbf{74.5} & 89.8 & \textbf{50.3} & \textbf{93.1} & \textbf{80.2} & \textbf{57.8} & \textbf{94.2} & 55.1 & \textbf{66.8} & \textbf{50.7} & 34.2 & \textbf{72.4} & \textbf{67.0} \\
    Ours   &                              & \textbf{91.6} & \textbf{55.7} & \textbf{90.5} & \textbf{43.9} & 49.2 & 60.4 & 64.4 & 58.1 & \textbf{91.0} & 49.8 & 92.8 & 76.0 & 37.9 & 92.3 & \textbf{68.9} & 63.9 & 15.9 & \textbf{52.7} & 60.7 & 64.0 \\
    \midrule
    FedAvg & \multirow{2}[2]{*}{Synscapes} & 89.5 & 51.0 & 81.3 & 40.7 & 38.9 & 57.5 & 54.5 & \textbf{68.1} & 89.8 & \textbf{51.1} & \textbf{93.8} & 71.8 & 33.2 & 89.9 & 39.3 & 25.1 & \textbf{6.5} & 34.1 & 62.0 & 56.8 \\
    Ours   &                               & \textbf{91.5} & \textbf{53.9} & \textbf{90.9} & \textbf{48.3} & \textbf{47.8} & \textbf{58.0} & \textbf{64.4} & 61.0 & \textbf{91.1} & 48.9 & 92.1 & \textbf{80.0} & \textbf{55.0} & \textbf{93.5} & \textbf{79.2} & \textbf{64.7} & 4.7 & \textbf{45.6} & \textbf{62.8} & \textbf{64.9} \\
    \midrule
    FedAvg & \multirow{2}[2]{*}{GTA5} & 84.7 & 39.2 & 89.6 & \textbf{45.1} & 49.3 & 53.8 & 63.3 & 49.7 & 90.0 & 43.9 & 90.6 & 78.8 & 49.5 & 92.1 & 52.0 & \textbf{71.6} & \textbf{52.0} & \textbf{35.4} & 46.7 & 64.1 \\
    Ours   &                            & \textbf{90.3} & \textbf{51.9} & \textbf{90.9} & 40.7 & \textbf{55.3} & \textbf{60.7} & \textbf{65.4} & \textbf{71.2} & \textbf{91.1} & \textbf{50.5} & \textbf{94.5} & 78.8 & \textbf{50.2} & \textbf{92.8} & \textbf{71.1} & 50.7 & 24.0 & 32.4 & \textbf{55.1} & \textbf{64.6} \\
    \bottomrule
    \end{tabular}%
  }
  \label{tab:per-class}
\end{table*}
\begin{table}[t]
  \caption{Performance of FedAvg and our method in the real-to-real federated domain generalization setting. CBM denotes the merged Cityscapes, BDD100K, and Mapillary datasets.}
  \centering
  \setlength{\tabcolsep}{6pt}
  \resizebox{0.48\textwidth}{!}{
    \begin{tabular}{c|cccc|ccc}
    \toprule
    \multirow{2}{*}{Method} & \multicolumn{4}{c|}{Dataset} & \multicolumn{3}{c}{mIoU} \\
    \cmidrule{2-8}
          & Client 1 & Client 2 & Client 3 & Server & IDD & ACDC & Avg \\
    \midrule
    Baseline 
          & - & - & - & CBM 
          & 62.7 & 70.7 & 66.7 \\
    \midrule

    FedAvg & \multirow{2}{*}{Cityscapes} 
           & \multirow{2}{*}{BDD100K} 
           & \multirow{2}{*}{Mapillary} 
           & \multirow{2}{*}{Urbansyn}
           & 53.6 & 49.6 & 51.6 \\
    FedS2R &  &  &  &  
           & \textbf{60.9} & \textbf{57.6} & \textbf{59.3} \\
    \midrule

    FedAvg & \multirow{2}{*}{Cityscapes} 
           & \multirow{2}{*}{BDD100K} 
           & \multirow{2}{*}{Mapillary} 
           & \multirow{2}{*}{Synscapes}
           & 49.4 & 46.4 & 47.9 \\
    FedS2R &  &  &  &  
           & \textbf{61.7} & \textbf{59.9} & \textbf{60.8} \\
    \midrule

    FedAvg & \multirow{2}{*}{Cityscapes} 
           & \multirow{2}{*}{BDD100K} 
           & \multirow{2}{*}{Mapillary} 
           & \multirow{2}{*}{Synthia}
           & 36.3 & 33.1 & 34.7 \\
    FedS2R &  &  &  &  
           & \textbf{59.8} & \textbf{59.2} & \textbf{59.5} \\
    \midrule

    FedAvg & \multirow{2}{*}{Cityscapes} 
           & \multirow{2}{*}{BDD100K} 
           & \multirow{2}{*}{Mapillary} 
           & \multirow{2}{*}{GTA5}
           & 57.4 & 50.4 & 53.9 \\
    FedS2R &  &  &  &  
           & \textbf{61.3} & \textbf{58.4} & \textbf{59.9} \\
    \bottomrule
    \end{tabular}
  }
  \label{tab:real2real}
\end{table}
\subsection{Datasets}
We establish our federated domain generalization setting using four synthetic datasets: GTA5 \cite{richter:2016GTA5}, Synscapes \cite{wrenninge:2018synscapes}, UrbanSyn \cite{gomez2025all}, and Synthia \cite{ros:2016}. These datasets employ different rendering technologies, leading to variations in image quality and coverage of distinct driving scenarios. Specifically, GTA5 contains 24,966 images at a resolution of $1914 \times 1052$; Synscapes includes 25,000 images at $1440 \times 720$; UrbanSyn comprises 7,539 images at $2048 \times 1024$; and Synthia provides 7,520 images rendered at $1280 \times 760$. The synthetic datasets are distributed across three clients, with each client retaining one dataset locally. The remaining dataset is allocated to the server and serves as the knowledge distillation set. To evaluate the generalization performance of our method, we adopt five real-world semantic segmentation benchmarks as unseen target domains: Cityscapes \cite{cordts:2016}, BDD100K \cite{yu:2020BDD}, Mapillary \cite{neuhold:2017mapillary}, IDD \cite{varma2019idd}, and ACDC \cite{sakaridis2021acdc}. Cityscapes consists of 2,975 training and 500 validation images at a resolution of $2048 \times 1024$. BDD100K includes 7,000 training and 1,000 validation images at $1280 \times 720$. Mapillary Vistas contains 18,000 training and 2,000 validation images with varying resolutions. IDD provides 6,993 finely annotated training images and 981 validation images at $1920 \times 1080$. ACDC comprises 1,600 training and 400 validation images at $1920 \times 1080$, captured under challenging conditions such as night, rain, and fog. All experiments are evaluated using the mean Intersection over Union (mIoU) computed across all semantic classes.
\subsection{Implementation details}
\noindent \textbf{Client models}. Each client employs the Mask2Former architecture with a Swin-Large backbone pre-trained on ImageNet-21K. Input images are randomly cropped to a resolution of $1024 \times 512$. Training is conducted using the AdamW optimizer with a learning rate of $1 \times 10^{-4}$, a batch size of 2, and 100 object queries over 90,000 iterations. All other hyperparameters follow the official Mask2Former configuration. 

\noindent \textbf{Global model}. The global model adopts the Mask2Former architecture with a Swin-Large backbone pre-trained on ImageNet-21K, and operates at an input resolution of $1024 \times 512$.To ensure compatibility with the concatenated outputs from multiple client models, the number of object queries in the global model is set to 300, matching the total number of queries aggregated from the $K=3$ client models (each with 100 queries). The optimization settings mirror those of the client models, employing the AdamW optimizer with a learning rate of $1 \times 10^{-4}$. The temperature parameter for KL divergence is set to $\tau=1.0$, and both loss terms are weighted equally with $\lambda_{\text{cls}} = \lambda_{\text{mask}} = 1.0$.

\noindent \textbf{Inconsistency-driven Augmentation.} We apply our inconsistency-driven augmentation strategy to eight dynamic foreground classes: \textit{person}, \textit{rider}, \textit{car}, \textit{truck}, \textit{bus}, \textit{train}, \textit{motorcycle}, and \textit{bicycle}. The inconsistency score threshold is set to 1.0. For each selected class, we generate 100 high-quality images using the Stable Diffusion XL (SDXL) pipeline \cite{podell2024sdxl}. Both the base model and the refiner are employed: the base model runs for 40 inference steps with a guidance scale of 7.5, followed by 20 refinement steps under the same guidance. To ensure semantic alignment with driving scenes, class-specific prompts are automatically constructed via ChatGPT. Negative prompts are fixed as ``cartoon, illustration, low quality, blurry, fantasy, surreal, painting'' to encourage photorealistic generation. All generated images are rendered at a resolution of $1024 \times 1024$ and incorporated into the server-side distillation dataset.
\subsection{Comparision with FedAvg}
\begin{figure}[t]
    \centering
    \includegraphics[width=1\linewidth]{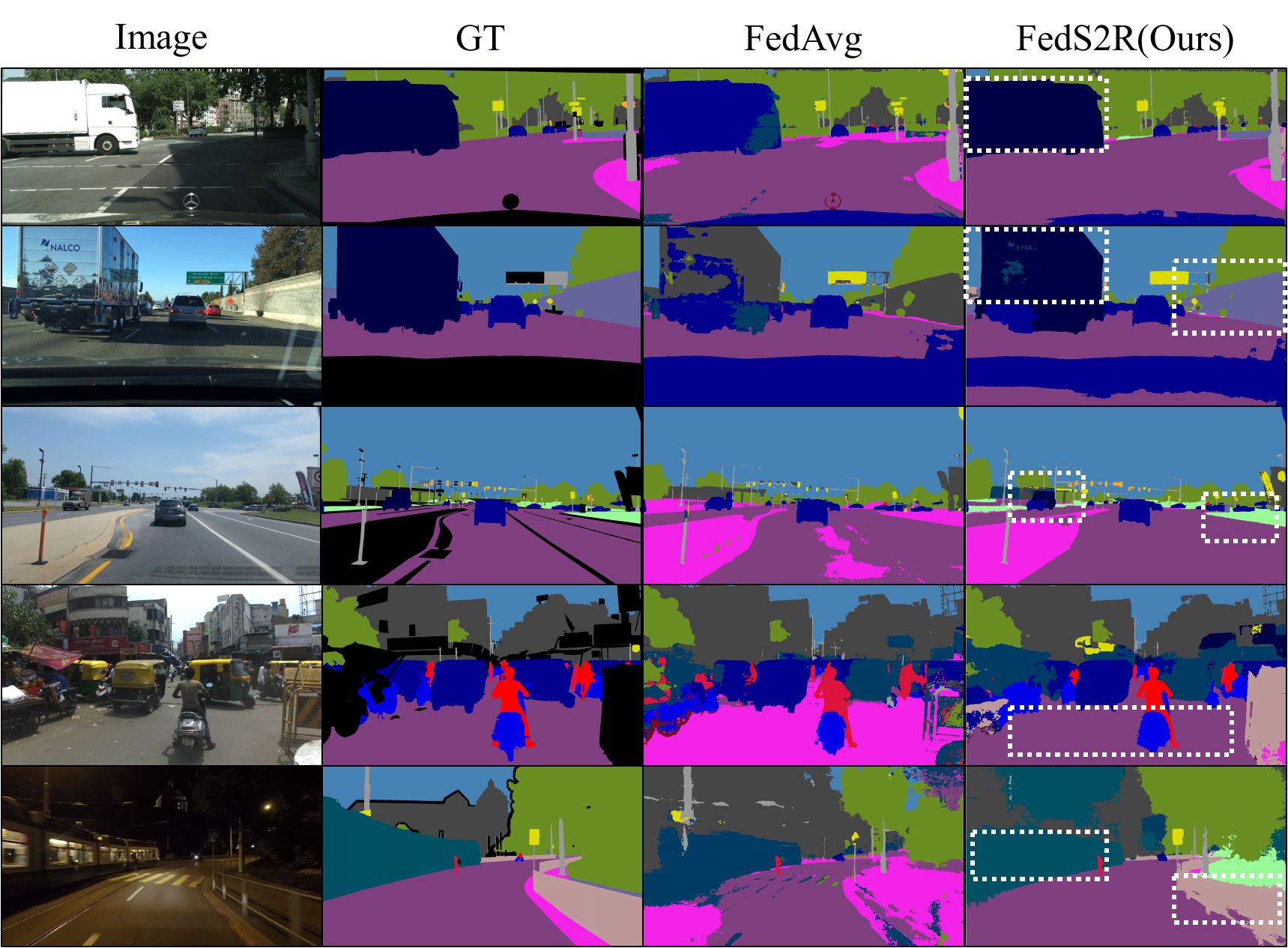}
    \caption{Qualitative comparisons between FedS2R and FedAvg under the configuration where Synthia serves as the server dataset. Rows 1–5 correspond to Cityscapes, BDD100K, Mapillary, IDD, and ACDC, respectively. FedS2R not only improves foreground classes such as \textit{train} and \textit{truck}, but also enhances background classes, including \textit{road}, \textit{terrain}, and \textit{fence}.}
    \label{fig:vis}
\end{figure}

Tab. \ref{tab:comparison} presents the performance of our method on five real domains compared with FedAvg under various client-server configurations. Since there is no existing work addressing this scenario, we compare our method with the representative baseline FedAvg. Specifically, we evaluate the performance of FedAvg under a one-shot federated learning setting for fair comparison. Across all configurations, our method consistently outperforms FedAvg in the averaged mIoU of all target domains. For example, when client datasets are GTA5, Synscapes, and UrbanSyn, and the server is Synthia, our method achieves an mIoU of 58.5, significantly outperforming FedAvg. This result is only 2.2 points below the upper-bound GSUS (model trained with simultaneous access to all datasets). Compared to baseline models trained individually on each client dataset, our method shows superior performance. Our method achieves 58.5 averaged mIoU, which is higher than any of the baselines: 53.7 for GTA5, 46.9 for Synscapes, and 51.9 for UrbanSyn.  Besides the averaged mIoU, our method also outperforms baseline models on target domains in specific cases. For example, it achieves 67.9 on Cityscapes, while baselines trained on GTA5, Synscapes, and UrbanSyn obtain 56.9, 54.0, and 63.9, respectively. This highlights the effectiveness of our framework in integrating knowledge across clients without access to client data. Besides quantitative results, we present visualisations in Fig. \ref{fig:vis} to illustrate the improvement of our framework.

Tab. \ref{tab:per-class} presents the per-class IoU results on the Cityscapes dataset across the same federated configurations of Tab. \ref{tab:comparison}. In most settings, our method consistently outperforms FedAvg. For Synthia as the server dataset, our method achieves a mIoU of 67.9, with a 22.9 points improvement over FedAvg. Notable gains are observed in several classes such as \textit{road} (92.3 vs 80.4), \textit{sidewalk} (56.6 vs 37.2), \textit{terrain} (51.9 vs 0), and \textit{train} (58.1 vs 0). An exception occurs in the UrbanSyn configuration; our method yields a mIoU of 64.0, which is lower than the 67.0 achieved by FedAvg. This performance gap is largely attributable to the high quality of the UrbanSyn dataset (note that the UrbanSyn baseline is by far the best baseline on Cityscapes), as well as the fact that FedAvg requires labeled server data during training, which is an advantage during training. Despite being behind, our method still improves the segmentation performance for several dynamic classes, such as \textit{truck} (68.9 vs 55.1) and \textit{motorcycle} (52.7 vs 34.2). Similarly, in the Synscapes configuration, our method improves the mIoU from 56.8 to 64.9, with particularly large gains in \textit{bus} (64.7 vs 25.1) and \textit{rider} (55.0 vs 33.2). When GTA5 serves as the server dataset, our method achieves a mIoU of 64.6, surpassing FedAvg by 0.5 points and offering improved segmentation for classes like \textit{pole} (60.7 vs 53.8) and \textit{truck} (71.1 vs 52.0). These results demonstrate that our framework consistently matches or exceeds the performance of FedAvg across a broad range of class types, including both background classes (e.g., \textit{building}, \textit{road}) and dynamic foreground classes (e.g., \textit{truck}, \textit{bus}, \textit{train}) without the necessity of labeled server data. 

In addition to the results with clients trained on synthetic datasets, we also evaluate the scenario where clients are trained on real-world data. Tab. \ref{tab:real2real} shows the performance under configurations in which the clients use real datasets while the server continues to rely on synthetic data. The results show that our method consistently outperforms FedAvg across all client–server combinations, indicating that our framework remains effective even when client domains shift from synthetic to real data. This highlights its robustness to heterogeneous client data while still leveraging a synthetic server dataset. 
\subsection{Ablation study}
\begin{table}[t]
  \caption{Ablation of components of FedS2R under the configuration of Synthia as the server dataset.}
  \centering
  \setlength{\tabcolsep}{12pt}
  \resizebox{0.5\textwidth}{!}{
    \begin{tabular}{cccccc}
    \toprule
    \multirow{2}[4]{*}{\shortstack{Feature\\fusion}} & \multicolumn{1}{c|}{\multirow{2}[4]{*}{\shortstack{Inconsistency\\augmentation}}} & \multicolumn{3}{c|}{Loss} & \multirow{2}[4]{*}{Avg mIoU} \\
    \cmidrule{3-5}          & \multicolumn{1}{c|}{} & \multicolumn{1}{c}{KL} & \multicolumn{1}{c}{Bce} & \multicolumn{1}{c|}{Dice} &  \\
    \midrule
      - & -   & \checkmark    & \checkmark    &  -  & 55.3  \\
     -  &  -  & \checkmark    &   - & \checkmark    & 26.0  \\
     -  &   - & \checkmark    & \checkmark    & \checkmark    & 56.4  \\
    \checkmark    & -   & \checkmark    & \checkmark    & \checkmark    & 57.1  \\
       - & \checkmark    & \checkmark    & \checkmark    & \checkmark    & 57.2  \\
    \checkmark    & \checkmark    & \checkmark    & \checkmark    & \checkmark    & 58.5  \\
    \bottomrule
    \end{tabular}%
    }
  \label{tab:ablation}%
\end{table}
Tab. \ref{tab:ablation} presents the ablation study evaluating the contribution of each component in our proposed framework under the configuration of Synthia as the server dataset. Using KL loss in combination with BCE loss yields an averaged mIoU of 55.3. However, replacing BCE loss with Dice loss results in a substantial drop to 26.0. This performance degradation suggests that Dice loss alone is insufficient for effectively supervising mask prediction in our federated setting. While Dice loss emphasizes region-level overlap, it provides limited guidance for learning accurate mask boundaries and shapes, which are more effectively captured by BCE loss. For feature fusion, which contributes an additional 0.7 points in averaged mIoU (from 56.4 to 57.1). This indicates that fusing features from client models facilitates the alignment of shared features across domains, which contributes to improved generalization. Adding inconsistency augmentation alone (without feature fusion) improves the averaged mIoU from 56.4 to 57.2 compared to the baseline. When combined with feature fusion, the full framework achieves the best performance (58.5). In summary, each component contributes meaningfully to the overall performance, and their combination leads to the best generalization across diverse and challenging target domains.

\section{Conclusion}
This paper proposes \framework, the first one-shot federated domain generalization framework for synthetic-to-real semantic segmentation in autonomous driving. \framework combines inconsistency-driven data augmentation and multi-client knowledge distillation with feature fusion. Extensive experiments on five real-world datasets, Cityscapes, BDD100K, Mapillary, IDD, and ACDC, show that the global model trained by our framework consistently outperforms all individual client models and performs only 2 mIoU points behind the model trained with simultaneous access to all client data. These results highlight the effectiveness and promise of federated domain generalization for semantic segmentation in autonomous driving.
\label{conclusion}

\bibliographystyle{./IEEEtran}
\bibliography{IEEEabrv}
\end{document}